\documentclass[10pt,twocolumn,letterpaper]{article}

\usepackage{cvpr}
\usepackage{times}
\usepackage{epsfig}
\usepackage{epstopdf}
\usepackage{graphicx}
\usepackage{amsmath}
\usepackage{amssymb}
\usepackage{verbatim}
\makeatletter
\newcommand*{\rom}[1]{\expandafter\@slowromancap\romannumeral #1@}
\makeatother
\providecommand{\e}[1]{\ensuremath{\times 10^{#1}}}

\usepackage[pagebackref=true,breaklinks=true,letterpaper=true,colorlinks,bookmarks=false]{hyperref}

 \cvprfinalcopy 

\def\cvprPaperID{66} 

\ifcvprfinal\fi
\begin{document}

\title{Uncovering Temporal Context for Video Question and Answering}

\author{Linchao Zhu$^{\S}$\hspace{1em}Zhongwen Xu$^\dag$\hspace{1em} Yi Yang$^\dag$\hspace{1em} Alexander G. Hauptmann$^\S$\\
	$^\S$SCS, Carnegie Mellon University\\
	$^\dag$QCIS, University of Technology Sydney\\
	{\tt\small \{zhulinchao7,zhongwen.s.xu,yee.i.yang\}@gmail.com} \hspace{1mm}  	 {\tt\small alex@cs.cmu.edu}
}

\maketitle
\begin{abstract}
   In this work, we introduce Video Question Answering in temporal domain to infer the past, describe the present and predict the future.
   We present an encoder-decoder approach using Recurrent Neural Networks to learn temporal structures of videos and
   introduce a dual-channel ranking loss to answer multiple-choice questions.
   We explore approaches for finer understanding of video content using question form of ``fill-in-the-blank'',
   and managed to collect 109,895 video clips with duration over 1,000 hours from TACoS, MPII-MD, MEDTest 14 datasets,
   while the corresponding 390,744 questions are generated from annotations.
   Extensive experiments demonstrate that our approach significantly outperforms the compared baselines.

\end{abstract}

\section{Introduction}

Current research into image analysis is gradually going beyond recognition~\cite{krizhevsky2012imagenet} and detection~\cite{girshick2014rich}. There are
 increasing interests in deeper understanding of visual content by jointly modeling image and natural language.
 As Convolutional Neural Networks (ConvNets) have raised the bar on image classification and detection tasks~\cite{girshick2014rich,DBLP:conf/icml/IoffeS15,szegedy2014going}, Recurrent Neural Networks (RNNs), particularly Long Short-Term Memory (LSTM)~\cite{lstm}, play a key role in visual description tasks, such as image captioning~\cite{donahue2014long,vinyals2014show,xu2015show}. As one step beyond image captioning, Image Question Answering (Image QA), which requires an extra layer of interaction between human and computers, have started to attract research attention very recently~\cite{antol2015vqa,gao2015you,malinowski2015ask}.

 In the area of video analysis, there are a few very recent systems proposed for video captioning~\cite{s2vt,yao2015capgenvid}. These methods have demonstrated promising
 performance in describing a video by a single short sentence. Similar as image captioning, video captioning may not be as intelligent as desired, especially when we only care about a particular part or object in the video~\cite{antol2015vqa}. In addition, it lacks the interaction between computers and the users~\cite{gao2015you}.

\begin{figure}
	\centering
	\includegraphics[width=1.0\linewidth,natwidth=313,natheight=278]{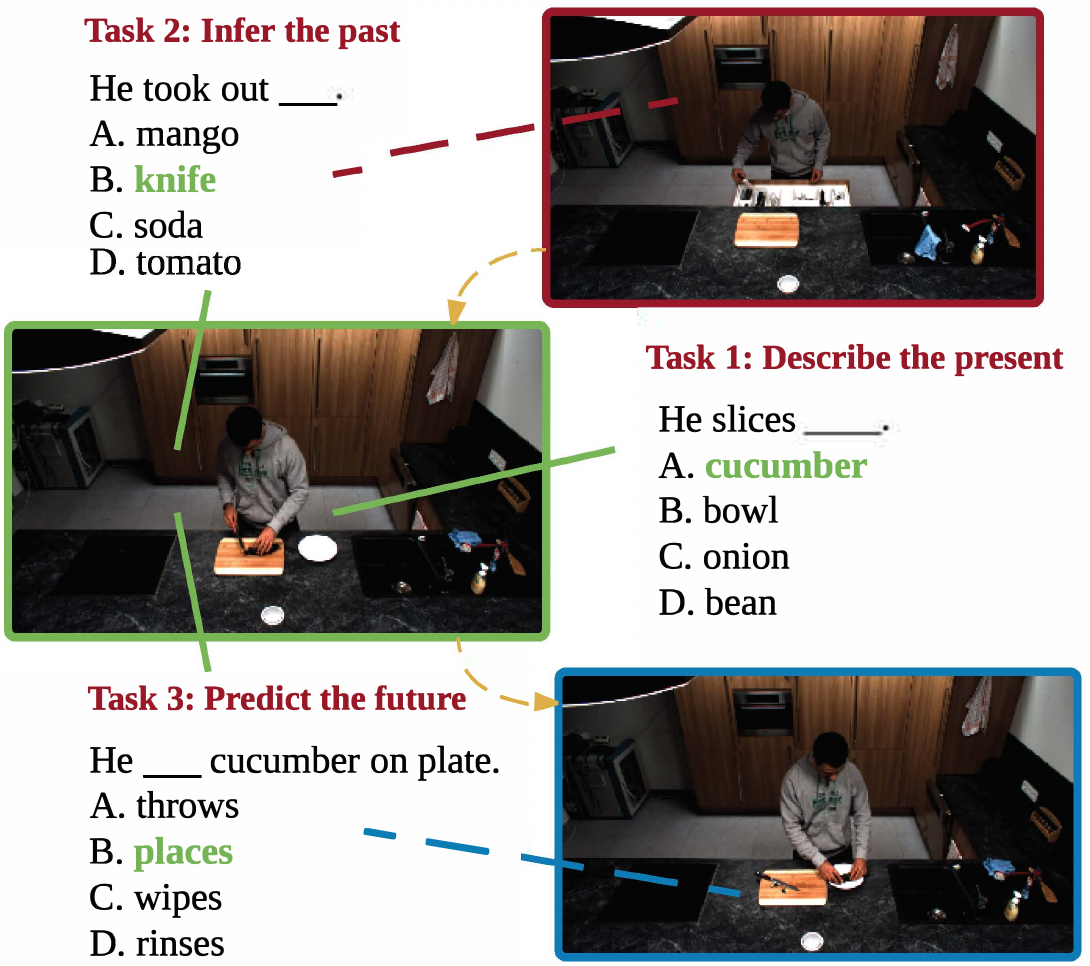}
	\caption{
		Questions and answers about the past, the present and the future. Our system includes three subtasks, which are inferring the \emph{past}, describing the \emph{present}, and predicting the \emph{future}, while \emph{only the current frames are observable}. Best viewed in color. 
	}
	\label{past_future_current_fig}
\end{figure}

In this paper, we focus on Video Question Answering (Video QA) in temporal domain, which has been largely unaddressed. Our Video QA consists of three subtasks. As shown in Figure \ref{past_future_current_fig}, if we see a man slicing cucumbers on a cutting board,
we can infer that he took out a knife \emph{previously}, and predict that he will put them on a plate \emph{afterwards}. The same as image QA, video QA requires finer understanding of videos and sentences than video captioning. Despite the success of these methods for video captioning~\cite{s2vt,yao2015capgenvid}, there are a few research challenges remain unsolved, which makes them not readily applicable to Video QA.

First, a Video QA system should explore more knowledge beyond just visual information and the coarse sentence annotations because it requires finer understanding of video content and questions. For the sake of video captioning, existing systems~\cite{s2vt,yao2015capgenvid} train LSTM models merely based on video content and the associated coarse sentence annotations. Because the size of description embedding matrix is very large but many words usually appears only a few (less than 10) times in all descriptions, the results overfit easily. Recent study~\cite{lin2015don} found that visual and textual information are mutually beneficial. We paved a new way of video QA, by appropriately integrating information of all types, including sentences, words, and visual cues, into a joint learning framework to maximize the mutual benefits, during which external knowledge bases (\eg BookCorpus~\cite{moviebook} and Google News~\cite{mikolov2013distributed}) can be readily incorporated. Because the external knowledge bases reflect the underlying correlations among related entities, our approach  is able to to better parse questions and video frames.

Second, a Video QA system should be capable of reasoning across video frames, including inferring the past, describing present, and predicting the future, which are strongly correlated. Very recently, Gated Recurrent Unit (GRU)~\cite{Cho_GRU} has demonstrated promising performance on sequence modeling tasks, partially because it has simpler neural structure than LSTM. On top of GRU, we propose an
\emph{encoder-decoder} approach with \emph{a dual-channel ranking loss} to learn three video representations, one for each Video QA subtasks,  \ie,  past inference, present description, and future prediction. One appealing feature of our approach is that, the encoder-decoder approach is able to model a wider range of temporal information, and the reduced number of weight parameters in GRU makes it more robust to overfitting in temporal modeling. 
Further, the approach avoids the needs of creating a large number of labels to train the sequence model by embedding visual feature to a semantic space.

Third, we should have a well-defined quantitative evaluation metric and datasets from different domains to track progress of this important research~\cite{antol2015vqa}. Manually providing groundtruth for  a large amount of videos is extremely human labor intensive. BLEU~\cite{papineni2002bleu} has been widely used as an evaluation metric for image captioning but a few research papers and competition reports have indicated that BLEU is not a reliable metric, and cannot reflect human judgment~\cite{kulkarni2011baby,cider_metric}. Following~\cite{lin2015don}, we evaluate our question and answering approach in the form of  ``fill-in-the-blank'' (FITB) from multiple choices. Under this theme, we managed to collect a new dataset consisting of over 100,000 real-world videos clips, and 400,000 designed questions with more than 1,000,000 candidate answers. This dataset will be released to the public, which can be used as benchmarks for this research. The main advantage is that  it is more convenient for quantitative evaluation than free-style question answering. Note that the difficulty of the questions can be controlled in designing candidate answers.

In this paper, we propose a new framework for video QA by carefully addressing the three aforementioned challenges. The rest of this paper is organized as follows. After introducing related works, we detail the large scale dataset we have collected for video QA tasks. We then present our approach of video temporal structure modeling and the dual-channel learning to rank method for question answering. Extensive experiments are conducted to validate our approach.

\section{Related Works}
\noindent\textbf{Neural networks in video analysis.}
Recently, many ConvNets based video feature learning methods have been proposed.
Simonyan and Zisserman \cite{simonyan2014two} propose to utilize optical flow images extracted from videos as the inputs to train ConvNets. Along with the ordinal RGB stream, two-stream ConvNets can achieve comparable performance with the state-of-the-art hand-crafted feature improved Dense Trajectories~\cite{wang2013action}. Tran~\etal~\cite{c3d} propose 3D ConvNets which capture temporal dynamics in video clips without the very time-consuming optical flow extraction procedure. Xu~\etal~\cite{xu2015discriminative} adapt the ConvNet frame-level features by VLAD pooling over the timestamps to generate video representation, which shows great advantages over the traditional average pooling. Recently, a general sequence to sequence framework \emph{encoder-decoder} was introduced by Sutskever~\etal~\cite{sutskever2014sequence}, which utilize a multilayered RNN to encode a sequence of input into one hidden state, then another RNN takes the encoded state as input and decode it into a sequence of output.
Ng~\etal~\cite{ng2015beyond} apply the encoder-decoder framework on large-scale video classification tasks. Srivastava~\etal~\cite{icml2015_srivastava} extend
this general model to learn features from consecutive frames and propose a composite model for unsupervised LSTM autoencoder.

\vspace{0.5em}\noindent\textbf{Bridging vision and language: captioning and question answering.}
There are increasing interests in the field of multimodal learning for bridging computer vision and
natural language understanding~\cite{donahue2014long,karpathy2015deep,s2vt,vinyals2014show,yao2015capgenvid}.
Captioning is one of the most popular tasks among them,
and Long Short-Term Memory (LSTM) is heavily used as a recurrent neural network language model to automatically generate a sequence of words conditioned on the visual features,
which is inspired by the general recurrent encoder-decoder  framework~\cite{sutskever2014sequence}.
However, captioning task only generates a generic description for entire image or video clip and it is difficult to evaluate the quality of generated sentences,
\ie, it's hard to judge one description is better than another one or not. In addition, it is still an open research problem of designing a proper metric for visual captioning,
which can reflect human judgment~\cite{compare_desc_metric,cider_metric}. In this work, we instead focus on more fine-grained description on video content,
and our method is simple to evaluate in multiple-choice form, \ie, correct or wrong answer.
Recently, a bunch
of QA datasets and systems have been developed on images \cite{antol2015vqa,gao2015you,malinowski2015ask,DBLP:journals/corr/RenKZ15}.
Ren~\etal~\cite{DBLP:journals/corr/RenKZ15} use a fixed-length answer with only one word for answering questions about images.
Gao~\etal~\cite{gao2015you} use a more complex dataset with free-style multilingual question-answer pairs,
however it is hard to evaluate the answers, usually human judges are required.
Lin~\etal~\cite{lin2015don} introduce an interesting multiple-choice fill-in-the-blank question answering task on abstract scene,
and Yu~\etal~\cite{madlibs} apply the task on natural images with various question templates.
Images are good sources for recognizing objects, however, a very important task, question answering on video content has not been explored yet.
Different from the still images, video analysis can utilize the temporal information across the frames, along with the object and scene information. The richer structural information in videos introduces potentially better understanding to the visual content while imposes challenges at the same time.

\vspace{0.5em}\noindent\textbf{Video Question Answering and temporal structure reasoning.} To the best of our knowledge,
the only work on video-based question answering is Tu~\etal~\cite{tu2014joint}, which builds a query answering system based
on a joint parsing graph from both text and videos. However, Tu~\etal~\cite{tu2014joint} constrain their model only on surveillance
videos of predefined structure, which cannot deal with open-ended questions. Differently, we cope with unconstrained videos of any kind, \eg, cooking scenario, DVD movies, web videos from YouTube, and develop a novel framework for visual understanding with dynamic temporal structure. In the aspect of temporal structure learning, action forecasting has been initially studied in~\cite{vondrick2015anticipating}.
To predict the potential actions, Vondrick~\etal~\cite{vondrick2015anticipating} propose to use a regression
loss built upon a ConvNet and forecast limited categories of actions and objects in a very short period, \eg, one second.
In contrast, we utilize a more flexible encoder-decoder framework, modeling a wider range of temporal information, and we mainly focus on multiple-choice question answering tasks in the temporal domain, which goes well beyond the standard visual recognition.



\section{Dataset Collection and Task Definitions}
The goal of our work is to present a Video QA system in temporal domain to infer the past,
describe the present and predict the future.
We first describe our dataset collection and the way to
automatically generate template questions in Section~\ref{dataset_design_qa_pair_generation}.
Task definitions and dataset analysis would be discussed in Section~\ref{dataset_tasks}.


\subsection{Dataset and QA Pairs Generation}
\label{dataset_design_qa_pair_generation}
We in total collect over 100,000 videos and 400,000 questions,
while QA pairs are generated from existing datasets in different domains, from cooking scenario, DVD movies, to web videos:
\begin{enumerate}
    \item \textbf{TACoS Multi-Level}~\cite{tacos:regnerietal:tacl}. TACoS dataset consists of 127 long videos with total 18,227 annotations in the \emph{cooking scenario}.
    It provides multiple sentence descriptions in fine-grained levels, \ie, for each short clip in the long videos.
\item \textbf{MPII-MD}~\cite{MPII-MD}. MPII-MD is collected from \emph{DVD movies} where descriptions are generated
          from movie scripts semi-automatically. The dataset contains 68,375 clips and one annotation on average is provided for each clip.
      \item \textbf{TRECVID MEDTest 14}~\cite{med14}. TRECVID MEDTest~14 is a complex event wild video dataset collected from \emph{web} hosting services such as YouTube. Videos in the dataset are about 1,300 hours in duration.
              The videos are untrimmed and the annotation is provided for each long video, which can be regarded as a coarse high-level summarization compared with TACoS and MPII-MD datasets.
\end{enumerate}
\label{dataset_preprocessing}

\begin{figure}[t]
\begin{center}
   \includegraphics[width=0.8\linewidth,natwidth=301,natheight=204]{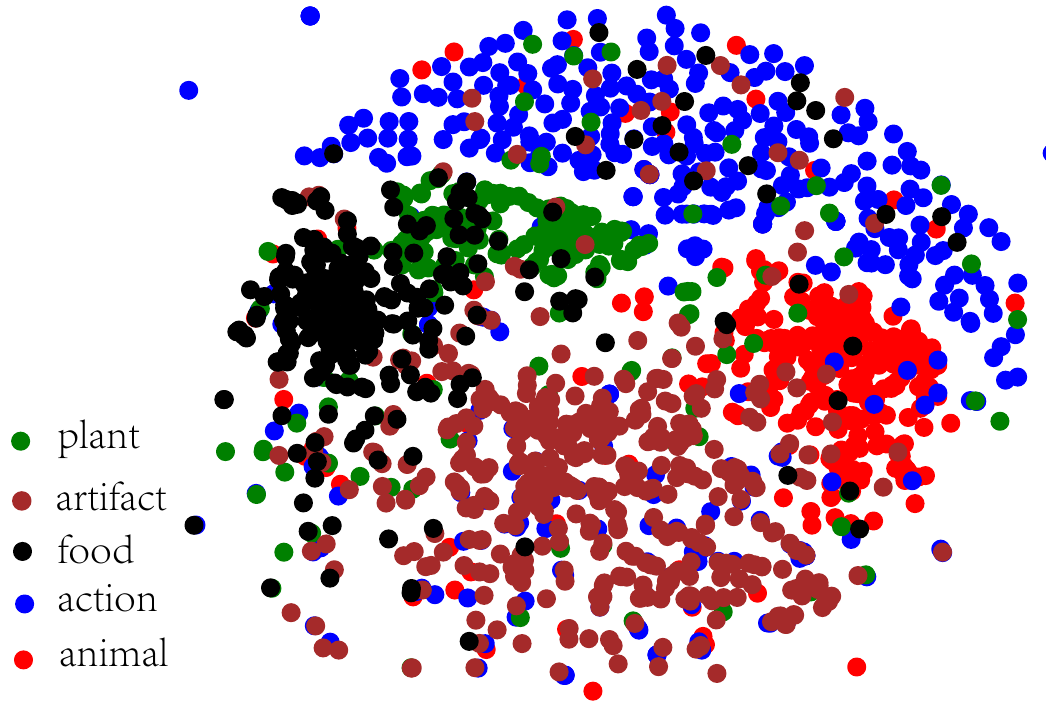}
\end{center}
   \caption{t-SNE visualization of word embeddings for each category learned from word2vec model. Best viewed in color.}
\label{visualize_of_categories}
\end{figure}

\noindent\textbf{Question templates generation.} We use the Stanford NLP Parser~\cite{klein2003accurate} to get syntactic structures of original video
descriptions. We divide the questions into three categories, nouns (objects like food, animals, plants),
verbs (actions) and phrases. Afterwards, question templates are generated from noun phrases (NP) and verb phrases (VP). During template generation,
we eliminate prepositional phrases as most of them are subjective.
We use WordNet\footnote{\url{https://wordnet.princeton.edu}} and NLTK\footnote{\url{http://www.nltk.org/}}
toolkits to identify word categories and choose a set of categories listed in Table \ref{dataset_num_count}.
We visualize the distribution of words in each category using t-SNE~\cite{van2008visualizing} in Figure~\ref{visualize_of_categories}. It shows that categories can be separated, where actions and objects have a clear margin.

\vspace{0.5em}
\noindent\textbf{Answer candidates generation.} We designed two different levels of difficulty in answering questions by altering candidate similarities.
For easy candidate pairs, we randomly choose three distractors within same category from the same dataset. Stop words like ``person'', ``man''
are filtered in advance and words with frequency less than 10 are filtered following the common practice.
As for hard pairs, based on the observations that video clips in the same dataset can be in totally different scenes, \eg, the MPII-MD dataset and the MEDTest 14 dataset, we select the hard negative candidates from similar descriptions.
In addition to the video datasets, we use description annotations from Flickr8K~\cite{hodosh2013framing}, Flickr30K~\cite{young2014image} and MS~COCO~\cite{lin2014microsoft} as description sources for similarity search.
We first parse the annotations using the way described above and gather about 8,000 phrases in total, resulting average length of 6.6 words per phrase.
After the preprocessing, we further filter the candidates using word2vec~\cite{mikolov2013distributed} to retrieve the nearest phrases in cosine distance. The phrase representation is generated by averaging the word vectors~\cite{DBLP:conf/icml/LebretPC15,lin2015don}. 

\begin{figure}[t]
\begin{center}
   \includegraphics[width=1.0\linewidth,natwidth=319,natheight=310]{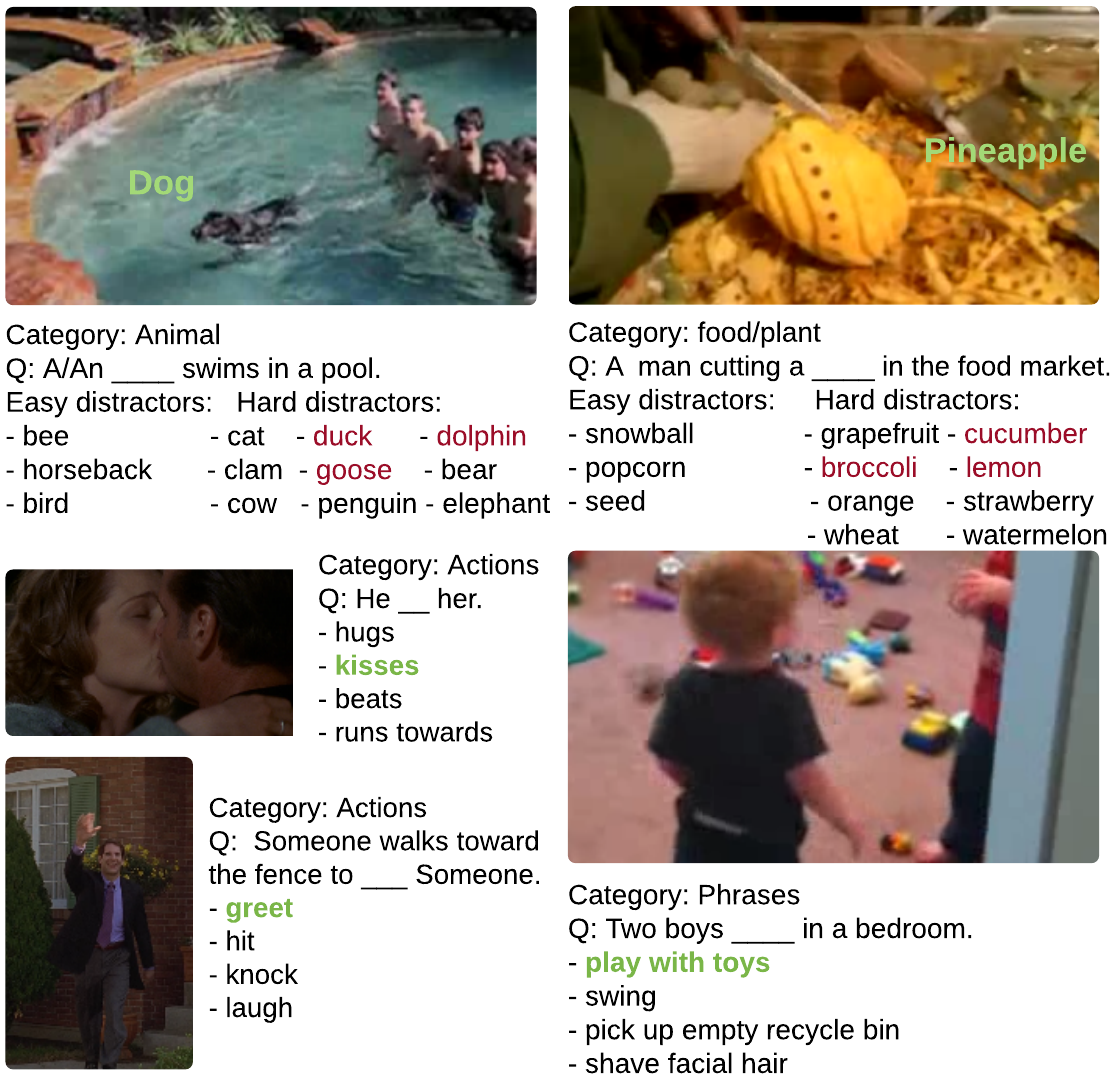}
\end{center}
   \caption{Examples of QA pairs for different categories and difficulty. Words colored in green are the correct answers, and difficult candidates are marked in red.}
\label{example_of_categories}
\end{figure}

As candidate answers might be ambiguous to the correct answer, we set a similarity threshold, and then select 10 of them as the final candidates.
We show examples of QA pairs in different categories and difficulty in Figure \ref{example_of_categories}.


\begin{table}
\footnotesize
\begin{center}
\tabcolsep=0.11cm
\begin{tabular}{c|ccccc}
Datasets & verbs & phrases & animals & food/plant & other objects \\
\hline\hline
TACoS      & 268        & 964          &  -           &      62         & 134      \\
MPII-MD    & 869        & 220          &  63          &      129        &     896 \\
MEDTest 14 & 671        & 418          &  98          &      174        &  726  \\
Combine all sources & 2,925       & 5,927         &  352         &  598            &  2,093 \\
\hline
\end{tabular}
\end{center}
\caption{
    List of categories and number of collected words in three datasets.
    Last rows shows the number of all words and phrases collected including those from
    image domains such as MS~COCO~\cite{lin2014microsoft}.
}
\label{dataset_num_count}
\end{table}

\subsection{Task Definitions}
\label{dataset_tasks}
Besides describing the current clip, we introduce another two tasks which are inferring the past and anticipating the future.
In the task of describing the present, we use all three datasets for evaluation.
As to the other two tasks which are past inferring and future predicting,
we perform experiments on TACoS and MPII-MD datasets only as they are annotated in fine-grained clips.
In these tasks, given a video clip, questions about the previous or next clip need to be answered.
Note that for tasks of describing the past and future, only the current clip is given and the model has to reason temporal structures based the given clip.
We restrict the past and future to be not too far away from the current clip and typically we choose the clip right before or after the given one,
where the time interval is less than 10 seconds.

For each task, we introduce two levels of questions.
For simplicity, we denote our tasks as \emph{Past-Easy},
\emph{Present-Easy},
\emph{Future-Easy},
\emph{Past-Hard},
\emph{Present-Hard}
and \emph{Future-Hard}.
We create three splits for each task and videos are divided into training, validation and testing sets.

\section{The Proposed Approach}
To answer questions about present, past and future, we first introduce
an encoder-decoder framework to represent context. We then map the visual representation to
semantic embedding space and learn to rank the correct answer with higher score.

\subsection{Learning to Represent Video Sequences}
\label{visual_learning_sec}
In this section, we describe our model of learning temporal context.
We present an encoder-decoder framework using Gated Recurrent Unit (GRU)~\cite{Cho_GRU}.
Compared with Long Short-Term Memory (LSTM) \cite{lstm}, GRU is conceptually simpler with
only two gates (update gates and reset gates) and no memory cells, while the performance on sequence modeling task
\cite{GRU_evaluation} is as good as LSTM. Note that we trained our model with LSTM as well,
but it performs worse than the one with GRU. With GRU, we can achieve mAP of 24.9\% on MEDTest 14 100Ex classification task, while we can only get 20.4\% with LSTM.
We suspect that it is because LSTM with more parameters is more prone to overfit than GRU.

\begin{figure}
    \centering
    \includegraphics[width=1.0\linewidth,natwidth=192,natheight=179]{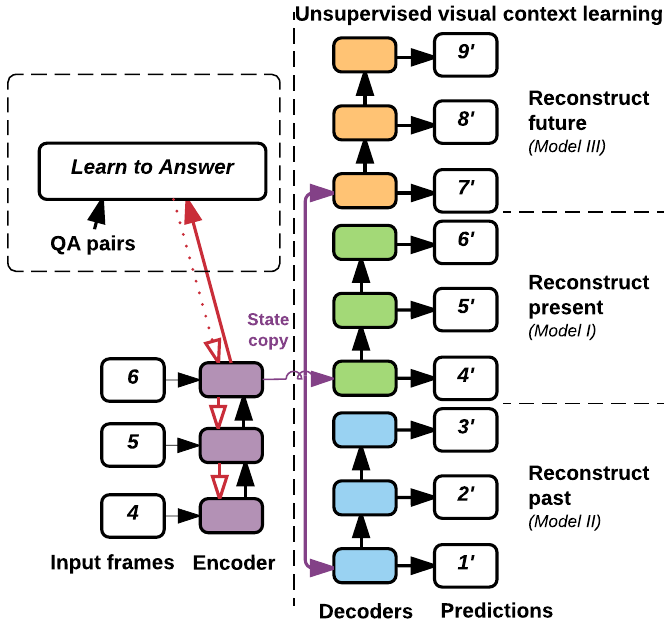}
    \caption{\footnotesize{
            The encoder-decoder model (right): encoder state of last time step is passed to three decoders for reconstruction.
            Learn to answer (left): learned to answer questions in a supervised way.
    }}
    \label{predict_and_remember_png}
\end{figure}

\noindent\textbf{Gated Recurrent Unit.} Denote $f_i^1,f_i^2, \ldots, f_i^N$ as the frames in a video $v_i$, where $N$ is
the number of frames sampled from the video. At each step $t$, the encoder generates
a hidden state $\mathbf{h}_i^t$, which can be regarded as the representation of sequence $f_i^1,f_i^2,\ldots,f_i^t$.
Thus the state of $\mathbf{h}_i^N$ encodes the whole sequence of frames.
States in GRU~\cite{Cho_GRU} are calculated as (dropping the video subscript $i$ for simplicity):
\begin{gather}
     \mathbf{r}^t = \sigma(W_{xr} \mathbf{x}^t + W_{hr} \mathbf{h}^{t-1}) \\
     \mathbf{z}^t = \sigma(W_{xz} \mathbf{x}^t + W_{hz} \mathbf{h}^{t-1}) \\
     \bar{\mathbf{h}^t} = \tanh(W_{x\bar{h}} \mathbf{x}^t + W_{h\bar{h}}(\mathbf{r}^t \odot \mathbf{h}^{t-1})) \\
     \mathbf{h}^t = (1 - \mathbf{z}^t) \odot \mathbf{h}^{t-1} + \mathbf{z}^t \odot \bar{\mathbf{h}^{t}}
\end{gather}
where $\mathbf{x}^t$ is the input, $\mathbf{r}^t$ is the reset gate, $\mathbf{z}^t$ is the update gate,
$\mathbf{h}^t$ is the proposed state and $\odot$ is element-wise multiplication.
For the decoder, we use the same architecture as the encoder, but its hidden state of $\mathbf{h}^0$ is initialized with the
hidden state of the last time step $N$ in the encoder.
Similar to \cite{icml2015_srivastava}, we construct our GRU encoder-decoder model (Figure~\ref{predict_and_remember_png}).
Besides reconstructing the input frames,
we also train another two models which are asked to reconstruct the future frames (Figure~\ref{predict_and_remember_png} top)
and past frames (Figure~\ref{predict_and_remember_png} bottom), respectively.
Our proposed models are capable of learning good features as
the network is optimized by minimizing the reconstruction error. In order to achieve good reconstruction,
representation passed to the decoder should retain high level abstraction of the target sequence.
Note that our three models are learned separately, where encoder and decoder weights are not shared across models of past, present and future.

\noindent\textbf{Training.} We first train the encoder-decoder models in an unsupervised way using videos collected from a subset of
MED dataset~\cite{med14} (exclude MEDTest~13 and MEDTest~14 videos) which consists of 35,805 videos with duration of over 1,300 hours.
The reason to choose MED dataset as a source for temporal context learning is that 
videos in MED dataset have much longer duration, containing complex and profound events, actions and objects for learning.
We collect data apart from our target task datasets as to learn more powerful model and practically,
it is difficult to train a model from scratch in such a small dataset like TACoS with only 127 cooking videos.
As frames in video are of high correlations in short range, we sample frames at the frame rate of 1 fps.
We use time span of 30 seconds and set the unroll length $T$ to 30 for the present model (Model~\rom{1}),
15 for both past model (Model~\rom{2}) and future model (Model~\rom{3}).

As for the input to GRU model, we use ConvNet features extracted from GoogLeNet~\cite{szegedy2014going}
with Batch Normalization~\cite{DBLP:conf/icml/IoffeS15} of dimension 1,024 which was trained from scratch
with ImageNet 2012 dataset~\cite{russakovsky2014imagenet} and we keep ConvNets part frozen during RNN training.

We now explain our network structures and training process in details.
As three models are trained with the same hyper-parameters, we take Model~\rom{1} as an example.
In our case, reconstruction error is measured by $\ell_2$ distance between predicted representation and the target sequence.
We reverse the target sequences in reconstruction scenario and
as indicated in \cite{sutskever2014sequence}, it reduces the path of the gradient flow.
We set the size of GRU units to 1,024 and two GRU layers are stacked.
Our decoders are conditioned on the inputs,
and we apply Dropout with rate 0.5 at connections between first GRU layer and second GRU layer as suggested by Zaremba~\etal~\cite{zaremba2014recurrent} to improve the generalization of the neural network.
We initialized $\mathbf{h}^0$ for encoder with zeros, while weights in input transformation layer are initialized with a uniform distribution in [-0.01, 0.01] and recurrent
weights are with uniform distribution in [-0.05, 0.05].
We set the mini-batch size to 64 and clip gradient element-wise at $1\e{-4}$. Frame sequences from different videos are sampled in each mini-batch.
The network is optimized by RMSprop~\cite{rmsprop}, which scales the gradient by a running average of gradient norm.
The model is trained by the Torch library~\cite{collobert2011torch7} on a single NVIDIA Tesla K20 GPU and it takes about one day for the models to converge and finish the training.

\noindent\textbf{Inference.} At inference time, we feed the ConvNet features extracted from GoogLeNet to the encoder, and obtain the video features from hidden states.
For each video clip, we initialized $\mathbf{h}^0$ to zeros, and pass the current hidden state to the next step until last input.
We then average hidden states at each time step as the final representation.


\subsection{Dual-Channel Learning to Rank}
\label{semantic_learning_sec}
We present the proposed dual-channel learning to rank algorithm
which jointly models two channels, \ie, word channel and sentence channel, for learning.
Kiros~\etal~\cite{DBLP:journals/corr/KirosZSZTUF15} recently propose the
skip-thought vectors to encode a sentence into a compact vector. The model uses an RNN encoder to encode a sentence and another two RNN decoders are asked to reconstruct the previous sentence and the next sentence.
It was trained using BookCorpus dataset \cite{moviebook} which consists of 11,038 books, 74,004,228 sentences and 984,846,357 words.
The skip-thought vectors model performs well on many different natural language processing (NLP) tasks.
We utilize the combine-skip model to encode sentences.
For more details, please refer to \cite{DBLP:journals/corr/KirosZSZTUF15}.

\label{ranking_loss_sec}

We first formulate the problem of multiple-choice question answering.
Given $N$ questions with blanks together with corresponding videos, and $K$ candidate answers
for each question, we denote each question as ${q}_i, i \in {1,\ldots,N}$, candidate answers
for question ${q}_i$ as ${p}_{ij}, j \in {1,\ldots,K}$ and the ground truth for question ${q}_i$ as
${p}_{i}'$ with index ${j}_i'$. For each question ${q}_i$, let ${s}_{ij}$ be the sentence
formed by filling the blank of question ${q}_i$ with candidate ${p}_{ij}$. For example,
filling in the template of ``A/An $\rule{0.8cm}{.15mm}$ swims in a pool'' shown in Figure \ref{example_of_categories} with candidate ``dog'',
we can form the sentence of ``A dog swims in a pool'',
and false description ``A horseback swims in a pool'' is generated with ``horseback''.

\begin{figure}
    \centering
    \includegraphics[width=1.0\linewidth,natwidth=201,natheight=151]{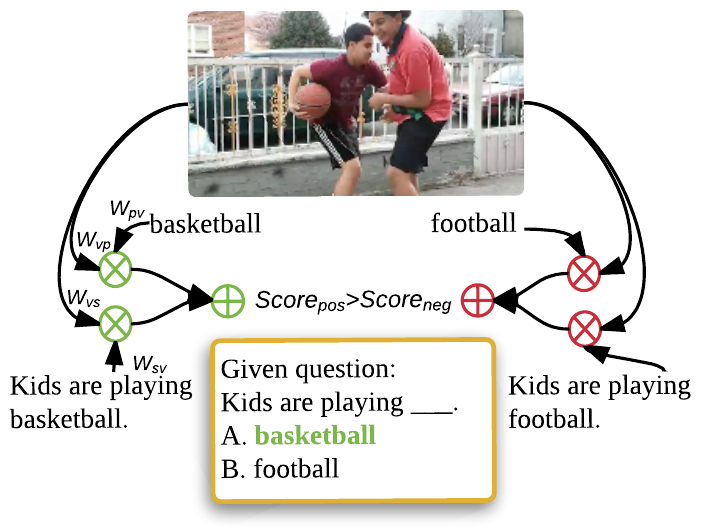}
    \caption{\footnotesize{
        Illustration of dual-channel learning to rank.
    }}
    \label{ranking_loss_png}
\end{figure}

Given ${q}_i$, we introduce a dual-channel ranking loss (also illustrated in Figure~\ref{ranking_loss_png}) that is trained to produce higher
similarity for the visual context and representation vector of the correct answer ${p}_i'$ than other distractors ${p}_{ij}, j \ne j'_i$.
We define our loss as:
\begin{equation}
    \min_{\boldsymbol\theta}  \sum_{\mathbf{v}}\sum_{j \in K, j \ne j'}\lambda \ell_{word} + (1 - \lambda)  \ell_{sent}, \lambda \in [0, 1],
\label{loss_function_equation}
\end{equation}
with
\begin{equation}\nonumber
\begin{aligned}
    & \ell_{word}=\max(0, \alpha - \mathbf{v_p}^T\mathbf{p}_{j'} + \mathbf{v_p}^T\mathbf{p}_{j}), \\
    & \ell_{sent}=\max(0, \beta - \mathbf{v_s}^T \mathbf{s}_{j'} + \mathbf{v_s}^T \mathbf{s}_{j}),
\end{aligned}
\end{equation}
where $\mathbf{v_p}=W_{vp}\mathbf{v}, \mathbf{v_s}=W_{vs}\mathbf{v}$ and $\mathbf{p}_j=W_{pv}\mathbf{y}_j, \mathbf{s}_j=W_{sv}\mathbf{z}_j$ (for simplicity we dropped subscript $i$).
$\mathbf{v}$ is the vector learning from our GRU encoder-decoder model for video clip $v_i$,
$\mathbf{y}_j$ is the average of
word2vec vectors for each word in candidate ${p}_{ij}$, $\mathbf{z}_j$ is the skip-thought vector for description $\mathbf{s}_{ij}$.
We constrain these feature representations to be in unit norm.
$\boldsymbol\theta$ denotes all the transformation parameters needed to learn in the model,
$W_{vs}$ and $W_{vp}$ are transformations that map visual representation to semantic joint space, while $W_{sv}$
and $W_{pv}$ transforms the semantic representation. Note that $W_{xx}$ can be a linear transformation or multi-layer neural networks with hidden units.

\noindent\textbf{Training.}
During training procedure, we sample false terms from negative candidates
and practically stop summing after first margin-violating term was found~\cite{frome2013devise}.
Empirically, we choose the sentence embedding dimension to be 500 and word embedding to be 300.
The model is trained by stochastic gradient descent (SGD) by simply setting the learning rate $\eta$ to be 0.01 and momentum with 0.9.
And in practice, we set the margin $\alpha$ and $\beta$ to 0.2, and $\lambda$ is cross-validated in held-out validation set.

\noindent\textbf{Inference.} We learned weight of transformations at training stage and at inference time, we calculate the following score for each candidate,
\begin{equation}
    score = \lambda \mathbf{v_p}^T\mathbf{p}_{j} + (1-\lambda)\mathbf{v_s}^T\mathbf{s}_{j},
    \label{score_equation}
\end{equation}
and the candidate with the highest score would be our answer.

\section{Experiments}
\subsection{Evaluation of Describing the Present}
In this section, we evaluate our model in the task of describing the present.
We first demonstrate the effectiveness of our ranking objective by comparing with CCA and then
conduct evaluation of dual-channel learning.

\noindent\textbf{Our dual-channel ranking method improves performance.}
We compare our dual-channel ranking approach with Canonical Correlation Analysis (CCA)
which computes the directions of maximal correlation between a pair of multi-dimensional variables. To learn CCA, we train two embedding layers separately.
The first CCA maps the sentence description to visual semantic joint-embedding space
and the second one maps the correct answer to the joint space. In order to answer multiple-choice questions, we embed each candidate and select the answer that is most similar to the video clip by Equation \ref{score_equation}.
We conduct cross-validation to choose the weight to combine two embeddings.

For both methods, we restrict the input features to be the same.
For visual representation, we average frame-level features extracted from the last fully connected layer of GoogLeNet.
For semantic representation, we use the same method described in Section \ref{semantic_learning_sec},
where sentences are encoded by skip-thought vectors, and word2vec is used for word representation.

Note that in CCA, the two embedding matrices are learned separately at training time while the weights of two embeddings are introduced at validation stage.
The method of late fusing sentence and word descriptions is different from our dual-channel ranking approach,
which integrates sentences and words representations during training time and learns to adjust embeddings accordingly. We demonstrate the effectiveness of our dual-channel ranking method
in Table~\ref{cca_table}.


\begin{table}
\footnotesize
\begin{center}
\begin{tabular}{lc|c|c}
Dataset    & Split & CCA & Our objective \\
\hline\hline
          & split 1   & 67.1\%  & \textbf{77.7\%} \\
TACoS     & split 2   & 64.9\%  & \textbf{78.3}\% \\
          & split 3   & 63.2\%  & \textbf{72.9}\% \\
          & mean      & 65.1\%  & \textbf{76.3}\% \\
\hline
          & split 1   & 36.2\%  & \textbf{73.4}\% \\
MPII-MD   & split 2   & 42.9\%  & \textbf{72.5}\% \\
          & split 3   & 45.7\%  & \textbf{69.9}\% \\
          & mean      & 41.6\%  & \textbf{72.0}\% \\
\hline
          & split 1   &  63.1\% & \textbf{81.2}\% \\
MEDTest 14 & split 2   &  62.8\% & \textbf{80.9}\% \\
          & split 3   &  63.6\% & \textbf{81.0}\% \\
          & mean      &  63.2\% & \textbf{81.0}\% \\
\hline
\end{tabular}
\end{center}
\caption{Comparison between CCA and our objective on \textit{Present-Easy} task.
The visual feature of averaging frame level 1,024 dimension representations from GoogLeNet is used for both approaches.
Our method outperforms CCA with a large margin.}
\label{cca_table}
\end{table}

%
As we can see, our objective outperforms CCA with a large margin. We believe it is because our objective function learns to integrate two representations,
while CCA uses a fixed embedding matrix during semantic weight learning.
Besides, CCA eliminates negative
terms during training, and as multiple-choice question-answering is required to select an answer from candidates at testing
time, ranking loss is more suitable for modeling the problem.

\begin{figure}[!h]
    \centering
    \includegraphics[height=0.6\linewidth]{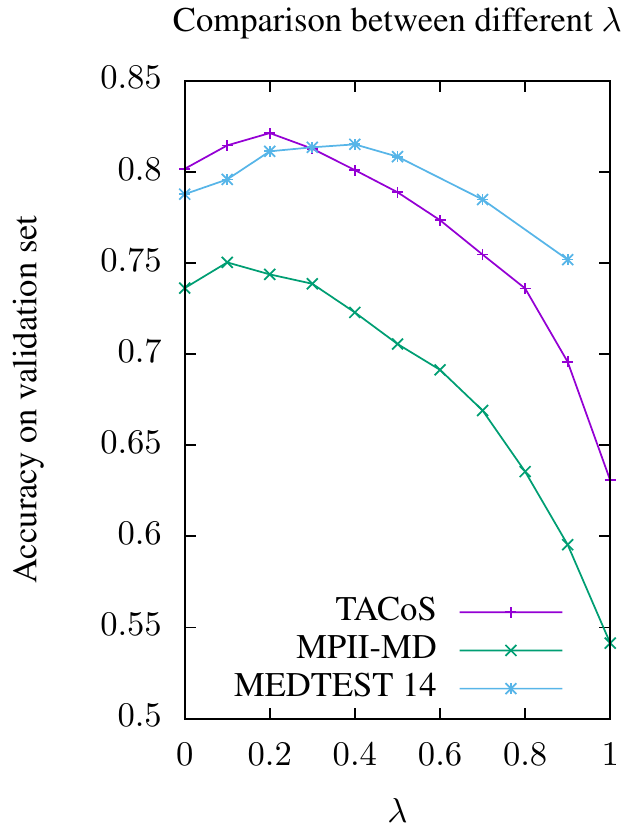}
%
\caption{The effectiveness of dual-channel learning to rank. We conduct experiment on \textit{Present-Easy} task to showcase. $\lambda = 0$ corresponds to using sentence channel only and $\lambda = 1$ corresponds to using word channel only.}
\label{sentence_phrase_example}
\end{figure}

\noindent\textbf{Evaluation of dual-channel learning.}
We then show
the effectiveness of using two channels for learning. 
The result of how integrating two representations influences the  performance is shown in Figure~\ref{sentence_phrase_example}.
As we can see, it is beneficial to integrate word representations during training, and
 sentences are weighted more than words. It is because our visual features represent more of global abstraction, which is corresponding to the sentence representation, 
while specific object features corresponding to the word representation haven't been considered in this work. We will explore this direction in details in the future works.

\begin{figure*}
\begin{center}
   \includegraphics[width=1.0\linewidth,natwidth=605,natheight=304]{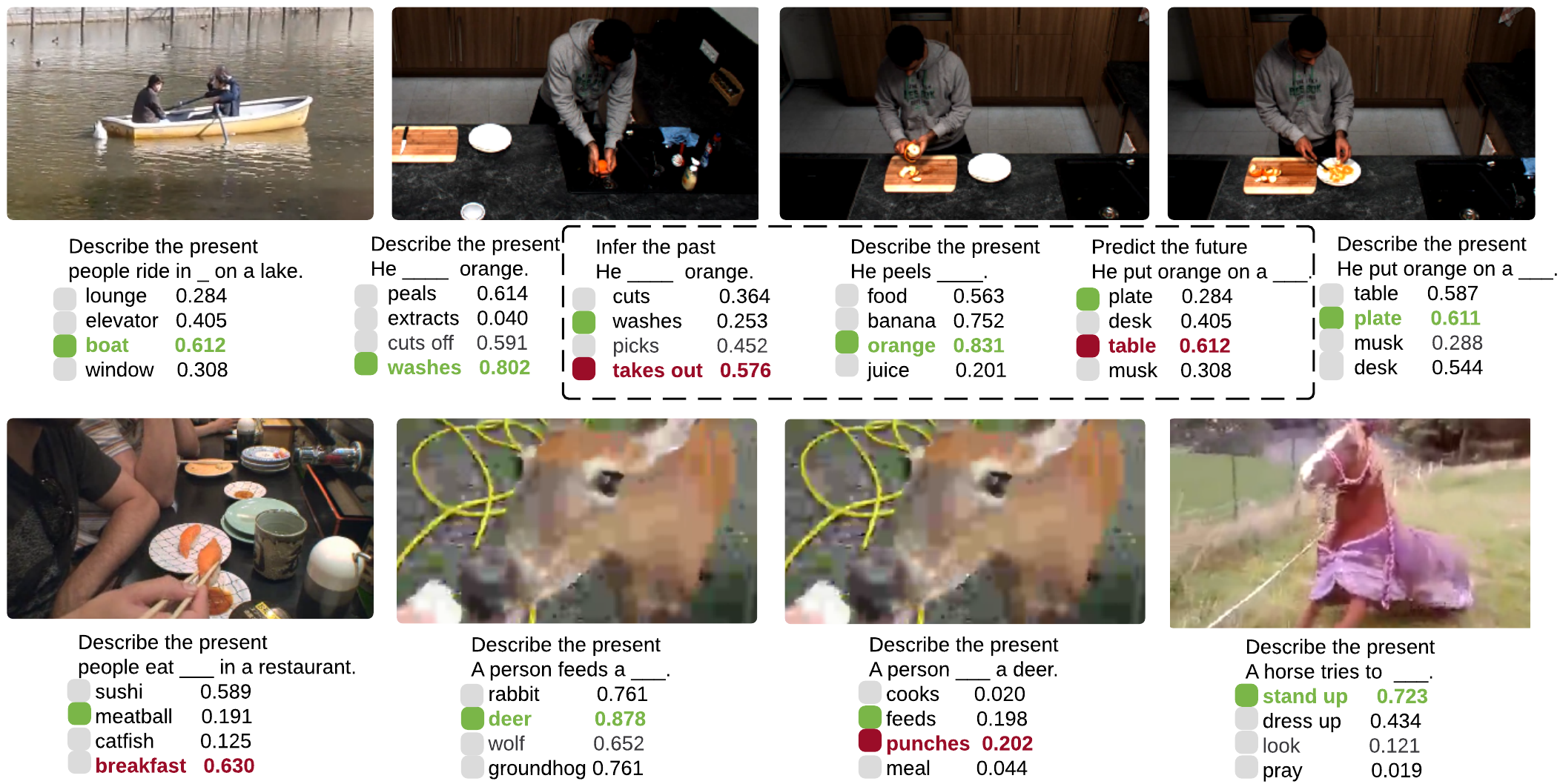}
\end{center}
   \caption{Example results obtained from our model. Each candidate has a score corresponding to a clip. Correct answers are marked in green while failed cases are in red.}
\label{full_model_result}
\end{figure*}

\noindent\textbf{Comparison between our GRU model and ConvNet model.}
\label{gru_cnn_comp_section}
To show the effectiveness of our encoder-decoder approach in modeling the present,
we compare our present model with a strong baseline
- averaging frame-level features from GoogLeNet. We compare two representations by placing the visual input to our dual-channel ranking objective with ConvNet model or our GRU model.

Note that the comparison is reasonable as both features are with same dimension of 1,024 and we use the same transformation layer and same hyper-parameters during training.
This result is shown in Table~\ref{cnn_past_future}. Detailed analysis will be discussed in next Section.

\begin{table}
\footnotesize
\begin{center}
\begin{tabular}{ll|c|c|c|c}
\hline
Level &  Dataset &  Split 1 & Split 2  & Split 3  & Mean    \\
\hline\hline
& TACoS          &  79.1\%  &  81.9\%  &  78.1\%  &  79.7\% \\
Easy & MPII-MD   &  75.5\%  &  74.6\%  &  72.4\%  &  74.2\% \\
& MEDTest 14     &  83.7\%  &  83.0\%  &  82.8\%  &  83.2\% \\
\hline
& TACoS          &  66.9\%  &  66.2\%  &  68.2\%  &  67.1\% \\
Hard & MPII-MD   &  47.4\%  &  49.0\%  &  48.3\%  &  48.2\% \\
& MEDTest 14     &  63.0\%  &  63.9\%  &  62.3\%  &  63.1\% \\
\hline
\end{tabular}
\end{center}
\caption{Results of our GRU model on the task of describing the present.}
\label{present_all_table}
\end{table}

\begin{table}
\footnotesize
\begin{center}
\begin{tabular}{lc|cc||cc}
\hline
&                          & \multicolumn{2}{c||}{Past} &  \multicolumn{2}{c}{Future} \\
Dataset &   Split                & Easy    & Hard   & Easy  & Hard \\
\hline\hline
           & split 1 & 78.1\%  &  65.8\% & 76.9\% &  66.1\% \\
TACoS      & split 2 & 78.3\%  &  64.4\% & 79.6\% &  65.8\% \\
           & split 3 & 78.5\%  &  63.9\% & 79.7\% &  69.9\% \\
\hline
           & split 1 & 72.4\%  &  47.0\% & 75.9\% &  47.1\% \\
MPII-MD    & split 2 & 72.0\%  &  47.0\% & 73.3\% &  48.8\% \\
           & split 3 & 72.0\%  &  46.9\% & 71.7\% &  48.1\% \\
\hline
\end{tabular}
\end{center}
\caption{Results of our GRU models on inferring past and predicting the future.}
\label{gru_all_table}
\end{table}

\begin{table*}
\footnotesize
\begin{center}
\begin{tabular}{lc|ccc||ccc||ccc}
\hline
&            &          & Past      &       &          & Present  &        &          & Future   &       \\
&            & ConvNets &  Ours     &Improv & ConvNets & Ours     &Improv  & ConvNets & Ours     &Improv \\
\hline\hline
& Easy& 74.8\%   & \textbf{78.3\%}    &3.5\%  &  76.3\%  & \textbf{79.7\%}   & 3.4\%  &  76.4\%  & \textbf{78.7\%}   & 2.3\% \\
TACoS  & Hard& 62.7\%   & \textbf{64.7\%}    &2.0\%  &  65.5\%  & \textbf{67.1\%}   & 1.6\%  &  64.5\%  & \textbf{67.3\%}   & 2.8\% \\
\hline
& Easy& 66.8\%   & \textbf{72.1\%}    &5.3\%  &  72.0\%  & \textbf{74.2\%}   & 2.2\%  &  68.7\%  & \textbf{73.6\%}   & 4.9\% \\
MPII-MD& Hard& 45.6\%   & \textbf{47.0\%}    &1.4\%  &  47.3\%  & \textbf{48.2\%}   & 0.9\%  &  46.9\%  & \textbf{48.0\%}  &1.1\% \\

\hline
\end{tabular}
\end{center}
\caption{Comparisons between ConvNets and our model for past, present and future modeling.}
\label{cnn_past_future}
\end{table*}

\subsection{Evaluation of Inferring the Past and Predicting Future}
We first show the results of our GRU models in all tasks. The results of describing the present is in Table~\ref{present_all_table},
while results of inferring the past and predicting the future are shown in Table~\ref{gru_all_table}.
We visualize part of the experiment results using our GRU models in Figure~\ref{full_model_result}
and some wrong answers are shown as well.


To demonstrate the effectiveness of our GRU models in modeling temporal structures, we conduct an interesting experiment which uses ConvNet features of the given clip to model
past and future directly. The results are shown in Table~\ref{cnn_past_future}. From the result, we have the following observations:

(1) \emph{GRU model outperforms ConvNet model in all cases, and relatively performs better than ConvNet in tasks of
inferring the past and predicting the future compared with describing the present.}
By comparisons of the performance among tasks, we find that our GRU model performs relatively better than ConvNets in tasks of
inferring the past and predicting the future, which shows the effectiveness of our GRU encoder-decoder framework in modeling temporal structures in videos.
As our GRU models are trained to reconstruct the past and future sequences, they can represent the past and future in a more reasonable way than the ConvNet models.
Our results also indicate the ability of our GRU models to capture
wider range of temporal information than ConvNet models. ConvNets trained from still frames can model temporal structures if objects in scene don't change too much in
short intervals (one example would be in Figure~\ref{past_future_current_fig}, ``cucumber'' occurs in both current and future clip). However,
when it comes to modeling longer sequences, ConvNets will fail to make predictions due to lack of context.


(2) \emph{Our model can achieve better results for future prediction than past inference.}
For future prediction, we feed input frames in the order of \emph{4, 5, 6} (Figure \ref{predict_and_remember_png})
and the decoder is asked to reconstruct frame in the order \emph{7, 8, 9}. As to past inferring, we feed the same input, but ask
the decoder to reconstruct target sequence of \emph{1, 2, 3}.
As the future prediction model has shorter term dependencies than past inferring model,
future prediction model can be easier to learn the temporal dependencies, which is consistent with the observations and hypothesis in~\cite{sutskever2014sequence}.



\subsection{Limitations and Future Work}
Although our results on question answering for video temporal context are encouraging,
our model has multiple limitations.
First, our model is only aware of context of at most 30 seconds (the longest unroll length).
One more flexible and promising approach would be
incorporating the attention mechanism~\cite{bahdanau2014neural} to learn longer sequences of context in videos.
Additionally, our model fails to answer questions about detailed objects sometimes, due to lack of local visual features, \ie, region-level, bounding boxes based representation.
We would like to integrate object detection ingredients to localize objects for better visual understanding. Lastly, we fixed sentence and word representation learning part in this work. Learning both
visual and language representations simultaneously is an open problem as indicated in~\cite{frome2013devise}.

\section{Conclusion}
Unlike video captioning tasks which generate a generic and single description for a video clip, we introduce an approach of temporal structure modeling for video question answering. 
We utilize an encoder-decoder model trained in an unsupervised way for visual context learning and propose a dual-channel learning to
ranking method to answer questions. The proposed method is capable of modeling video temporal structure in a longer time range.
We evaluate our approach on three datasets which have a large number of videos.
The new approach outperforms the compared baselines,  and achieves encouraging question answering results. 
{\small
\bibliographystyle{ieee}
\bibliography{qa}

\begin{thebibliography}{10}\itemsep=-1pt

\bibitem{med14}
{TRECVID} {MED} 14.
\newblock \url{http://nist.gov/itl/iad/mig/med14.cfm}, 2014.

\bibitem{antol2015vqa}
S.~Antol, A.~Agrawal, J.~Lu, M.~Mitchell, D.~Batra, C.~L. Zitnick, and
  D.~Parikh.
\newblock {VQA}: Visual question answering.
\newblock In {\em ICCV}, 2015.

\bibitem{bahdanau2014neural}
D.~Bahdanau, K.~Cho, and Y.~Bengio.
\newblock Neural machine translation by jointly learning to align and
  translate.
\newblock In {\em ICLR}, 2015.

\bibitem{Cho_GRU}
K.~Cho, B.~van Merrienboer, C.~Gulcehre, F.~Bougares, H.~Schwenk, and
  Y.~Bengio.
\newblock Learning phrase representations using {RNN} encoder-decoder for
  statistical machine translation.
\newblock In {\em EMNLP}, 2015.

\bibitem{GRU_evaluation}
J.~Chung, C.~Gulcehre, K.~Cho, and Y.~Bengio.
\newblock Empirical evaluation of gated recurrent neural networks on sequence
  modeling.
\newblock {\em arXiv preprint arXiv:1412.3555}, 2014.

\bibitem{collobert2011torch7}
R.~Collobert, K.~Kavukcuoglu, and C.~Farabet.
\newblock Torch7: A matlab-like environment for machine learning.
\newblock In {\em BigLearn, NIPS Workshop}, 2011.

\bibitem{donahue2014long}
J.~Donahue, L.~A. Hendricks, S.~Guadarrama, M.~Rohrbach, S.~Venugopalan,
  K.~Saenko, and T.~Darrell.
\newblock Long-term recurrent convolutional networks for visual recognition and
  description.
\newblock In {\em CVPR}, 2015.

\bibitem{compare_desc_metric}
D.~Elliott and F.~Keller.
\newblock Comparing automatic evaluation measures for image description.
\newblock In {\em ACL}, 2014.

\bibitem{frome2013devise}
A.~Frome, G.~S. Corrado, J.~Shlens, S.~Bengio, J.~Dean, M.~A. Ranzato, and
  T.~Mikolov.
\newblock {DeViSE}: A deep visual-semantic embedding model.
\newblock In {\em NIPS}, 2013.

\bibitem{gao2015you}
H.~Gao, J.~Mao, J.~Zhou, Z.~Huang, L.~Wang, and W.~Xu.
\newblock Are you talking to a machine? {Dataset} and methods for multilingual
  image question answering.
\newblock In {\em NIPS}, 2015.

\bibitem{girshick2014rich}
R.~Girshick, J.~Donahue, T.~Darrell, and J.~Malik.
\newblock Rich feature hierarchies for accurate object detection and semantic
  segmentation.
\newblock In {\em CVPR}, 2014.

\bibitem{lstm}
S.~Hochreiter and J.~Schmidhuber.
\newblock Long short-term memory.
\newblock {\em Neural computation}, 9(8):1735--1780, 1997.

\bibitem{hodosh2013framing}
M.~Hodosh, P.~Young, and J.~Hockenmaier.
\newblock Framing image description as a ranking task: Data, models and
  evaluation metrics.
\newblock {\em JAIR}, pages 853--899, 2013.

\bibitem{DBLP:conf/icml/IoffeS15}
S.~Ioffe and C.~Szegedy.
\newblock Batch normalization: Accelerating deep network training by reducing
  internal covariate shift.
\newblock In {\em ICML}, 2015.

\bibitem{karpathy2015deep}
A.~Karpathy and L.~Fei-Fei.
\newblock Deep visual-semantic alignments for generating image descriptions.
\newblock In {\em CVPR}, 2015.

\bibitem{DBLP:journals/corr/KirosZSZTUF15}
R.~Kiros, Y.~Zhu, R.~Salakhutdinov, R.~S. Zemel, A.~Torralba, R.~Urtasun, and
  S.~Fidler.
\newblock Skip-thought vectors.
\newblock In {\em NIPS}, 2015.

\bibitem{klein2003accurate}
D.~Klein and C.~D. Manning.
\newblock Accurate unlexicalized parsing.
\newblock In {\em ACL}, 2003.

\bibitem{krizhevsky2012imagenet}
A.~Krizhevsky, I.~Sutskever, and G.~E. Hinton.
\newblock {ImageNet} classification with deep convolutional neural networks.
\newblock In {\em NIPS}, 2012.

\bibitem{kulkarni2011baby}
G.~Kulkarni, V.~Premraj, S.~Dhar, S.~Li, Y.~Choi, A.~C. Berg, and T.~L. Berg.
\newblock Baby talk: Understanding and generating image descriptions.
\newblock In {\em CVPR}, 2011.

\bibitem{DBLP:conf/icml/LebretPC15}
R.~Lebret, P.~O. Pinheiro, and R.~Collobert.
\newblock Phrase-based image captioning.
\newblock In {\em ICML}, 2015.

\bibitem{lin2014microsoft}
T.-Y. Lin, M.~Maire, S.~Belongie, J.~Hays, P.~Perona, D.~Ramanan,
  P.~Doll{\'a}r, and C.~L. Zitnick.
\newblock Microsoft {COCO}: Common objects in context.
\newblock In {\em ECCV}. 2014.

\bibitem{lin2015don}
X.~Lin and D.~Parikh.
\newblock Don't just listen, use your imagination: Leveraging visual common
  sense for non-visual tasks.
\newblock In {\em CVPR}, 2015.

\bibitem{malinowski2015ask}
M.~Malinowski, M.~Rohrbach, and M.~Fritz.
\newblock Ask your neurons: A neural-based approach to answering questions
  about images.
\newblock In {\em ICCV}, 2015.

\bibitem{mikolov2013distributed}
T.~Mikolov, I.~Sutskever, K.~Chen, G.~S. Corrado, and J.~Dean.
\newblock Distributed representations of words and phrases and their
  compositionality.
\newblock In {\em NIPS}, 2013.

\bibitem{ng2015beyond}
J.~Y.-H. Ng, M.~Hausknecht, S.~Vijayanarasimhan, O.~Vinyals, R.~Monga, and
  G.~Toderici.
\newblock Beyond short snippets: Deep networks for video classification.
\newblock In {\em CVPR}, 2015.

\bibitem{papineni2002bleu}
K.~Papineni, S.~Roukos, T.~Ward, and W.-J. Zhu.
\newblock {BLEU}: a method for automatic evaluation of machine translation.
\newblock In {\em ACL}, 2002.

\bibitem{tacos:regnerietal:tacl}
M.~Regneri, M.~Rohrbach, D.~Wetzel, S.~Thater, B.~Schiele, and M.~Pinkal.
\newblock Grounding action descriptions in videos.
\newblock {\em TACL}, 1:25--36, 2013.

\bibitem{DBLP:journals/corr/RenKZ15}
M.~Ren, R.~Kiros, and R.~S. Zemel.
\newblock Image question answering: {A} visual semantic embedding model and a
  new dataset.
\newblock In {\em NIPS}, 2015.

\bibitem{MPII-MD}
A.~Rohrbach, M.~Rohrbach, N.~Tandon, and B.~Schiele.
\newblock A dataset for movie description.
\newblock In {\em CVPR}, 2015.

\bibitem{russakovsky2014imagenet}
O.~Russakovsky, J.~Deng, H.~Su, J.~Krause, S.~Satheesh, S.~Ma, Z.~Huang,
  A.~Karpathy, et~al.
\newblock {ImageNet} large scale visual recognition challenge.
\newblock {\em IJCV}, pages 1--42, 2014.

\bibitem{simonyan2014two}
K.~Simonyan and A.~Zisserman.
\newblock Two-stream convolutional networks for action recognition in videos.
\newblock In {\em NIPS}, 2014.

\bibitem{icml2015_srivastava}
N.~Srivastava, E.~Mansimov, and R.~Salakhudinov.
\newblock Unsupervised learning of video representations using {LSTMs}.
\newblock In {\em ICML}, 2015.

\bibitem{sutskever2014sequence}
I.~Sutskever, O.~Vinyals, and Q.~V. Le.
\newblock Sequence to sequence learning with neural networks.
\newblock In {\em NIPS}, 2014.

\bibitem{szegedy2014going}
C.~Szegedy, W.~Liu, Y.~Jia, P.~Sermanet, S.~Reed, D.~Anguelov, D.~Erhan,
  V.~Vanhoucke, and A.~Rabinovich.
\newblock Going deeper with convolutions.
\newblock In {\em CVPR}, 2015.

\bibitem{rmsprop}
T.~Tieleman and G.~Hinton.
\newblock Lecture 6.5-{RMSprop}: Divide the gradient by a running average of
  its recent magnitude.
\newblock 2012.

\bibitem{c3d}
D.~Tran, L.~Bourdev, R.~Fergus, L.~Torresani, and M.~Paluri.
\newblock Learning spatiotemporal features with {3D} convolutional networks.
\newblock In {\em ICCV}, 2015.

\bibitem{tu2014joint}
K.~Tu, M.~Meng, M.~W. Lee, T.~E. Choe, and S.-C. Zhu.
\newblock Joint video and text parsing for understanding events and answering
  queries.
\newblock {\em MultiMedia, IEEE}, 21(2):42--70, 2014.

\bibitem{van2008visualizing}
L.~Van~der Maaten and G.~Hinton.
\newblock Visualizing data using t-{SNE}.
\newblock {\em JMLR}, 9(2579-2605):85, 2008.

\bibitem{cider_metric}
R.~Vedantam, C.~L. Zitnick, and D.~Parikh.
\newblock {CIDEr}: Consensus-based image description evaluation.
\newblock In {\em CVPR}, 2015.

\bibitem{s2vt}
S.~Venugopalan, M.~Rohrbach, J.~Donahue, R.~Mooney, T.~Darrell, and K.~Saenko.
\newblock Sequence to sequence -- video to text.
\newblock In {\em ICCV}, 2015.

\bibitem{vinyals2014show}
O.~Vinyals, A.~Toshev, S.~Bengio, and D.~Erhan.
\newblock Show and tell: A neural image caption generator.
\newblock In {\em CVPR}, 2015.

\bibitem{vondrick2015anticipating}
C.~Vondrick, H.~Pirsiavash, and A.~Torralba.
\newblock Anticipating the future by watching unlabeled video.
\newblock {\em arXiv preprint arXiv:1504.08023}, 2015.

\bibitem{wang2013action}
H.~Wang and C.~Schmid.
\newblock Action recognition with improved trajectories.
\newblock In {\em ICCV}, 2013.

\bibitem{xu2015show}
K.~Xu, J.~Ba, R.~Kiros, A.~Courville, R.~Salakhutdinov, R.~Zemel, and
  Y.~Bengio.
\newblock Show, attend and tell: Neural image caption generation with visual
  attention.
\newblock In {\em ICML}, 2015.

\bibitem{xu2015discriminative}
Z.~Xu, Y.~Yang, and A.~G. Hauptmann.
\newblock A discriminative {CNN} video representation for event detection.
\newblock In {\em CVPR}, 2015.

\bibitem{yao2015capgenvid}
L.~Yao, A.~Torabi, K.~Cho, N.~Ballas, C.~Pal, H.~Larochelle, and A.~Courville.
\newblock Describing videos by exploiting temporal structure.
\newblock In {\em ICCV}, 2015.

\bibitem{young2014image}
P.~Young, A.~Lai, M.~Hodosh, and J.~Hockenmaier.
\newblock From image descriptions to visual denotations: New similarity metrics
  for semantic inference over event descriptions.
\newblock {\em TACL}, 2:67--78, 2014.

\bibitem{madlibs}
L.~Yu, E.~Park, A.~C. Berg, and T.~L. Berg.
\newblock Visual {Madlibs}: Fill in the blank image generation and question
  answering.
\newblock In {\em ICCV}, 2015.

\bibitem{zaremba2014recurrent}
W.~Zaremba, I.~Sutskever, and O.~Vinyals.
\newblock Recurrent neural network regularization.
\newblock {\em arXiv preprint arXiv:1409.2329}, 2014.

\bibitem{moviebook}
Y.~Zhu, R.~Kiros, R.~Zemel, R.~Salakhutdinov, R.~Urtasun, A.~Torralba, and
  S.~Fidler.
\newblock Aligning books and movies: Towards story-like visual explanations by
  watching movies and reading books.
\newblock In {\em ICCV}, 2015.

\end{thebibliography}
}

\end{document}